\documentclass[letterpaper, 10 pt, conference]{ieeeconf}  % Comment this line out if you need a4paper

\IEEEoverridecommandlockouts                              % This command is only needed if 
\usepackage{graphics} % for pdf, bitmapped graphics files
\usepackage{epsfig} % for postscript graphics files
\usepackage{mathptmx} % assumes new font selection scheme installed
\usepackage{times} % assumes new font selection scheme installed
\usepackage{amsmath} % assumes amsmath package installed
\usepackage{amssymb}  % assumes amsmath package installed
\usepackage{xcolor}
\usepackage{multicol}
\usepackage{multirow}
\usepackage{colortbl}
\usepackage{lipsum}
\usepackage{hyperref}
\DeclareMathAlphabet{\mathcal}{OMS}{cmsy}{m}{n}

\usepackage{ragged2e}

\title{\LARGE \bf
Enhancing Flexibility and Adaptability in Conjoined Human-Robot Industrial Tasks with a Minimalist Physical Interface
}

\author{Juan M. Gandarias$^{*+}$, Pietro Balatti$^{+}$, Edoardo Lamon, Marta Lorenzini, and Arash Ajoudani% <-this % stops a space
\thanks{This work was supported in part by the ERC-StG Ergo-Lean (Grant Agreement No.850932), in part by the European Union’s Horizon 2020 research and innovation programme under Grant Agreement No. 101016007 (CONCERT), in part by the  BRISK project ``Piano di attività 2019/2021 – Ricerca scientifica (bando BRiC 2019)''.}% <-this % stops a space
\thanks{The authors are with the HRI$^{2}$ Lab, Istituto Italiano di Tecnologia, Genoa, Italy.
{\tt\small \{juan.gandarias, pietro.balatti, edoardo.lamon, marta.lorenzini, arash.ajoudani\}@iit.it}}
\thanks{$^*$ Corresponding author.}
\thanks{$^{+}$ Contributed equally to this work.}
}

\begin{document}

\maketitle
\thispagestyle{empty}
\pagestyle{empty}

%%%%%%%%%%%%%%%%%%%%%%%%%%%%%%%%%%%%%%%%%%%%%%%%%%%%%%%%%%%%%%%%%%%%%%%%%%%%%%%%
\begin{abstract}
This paper presents a physical interface for collaborative mobile manipulators in industrial manufacturing and logistics applications. 
The proposed work builds on our earlier MOCA-MAN interface, through which an operator could be physically coupled to a mobile manipulator to be assisted in performing daily activities. The previous interface was based on a magnetic clamp attached to one arm of the user for the coupling stage, and a bracelet based on EMG sensors on the other arm for human-robot communication via gestures. The new interface instead presents the following additions: i) An industrial-like design that allows the worker to couple/decouple easily and to operate mobile manipulators locally; ii) A simplistic communication channel via a simple buttons board that allows controlling the robot with one hand only; iii) The interface offers enhanced loco-manipulation capabilities that do not compromise the worker mobility. In addition, an experimental evaluation with six human subjects is carried out to analyze the enhanced locomotion and flexibility of the proposed interface in terms of mobility constraint, usability, and physical load reduction.
\end{abstract}

%%%%%%%%%%%%%%%%%%%%%%%%%%%%%%%%%%%%%%%%%%%%%%%%%%%%%%%%%%%%%%%%%%%%%%%%%%%%%%%%
\section{Introduction}

% Motivation - collaborative robotic systems for industry and other applications
The introduction of robotic technologies in industrial environments has been one of the breakthroughs in manufacturing and logistics applications. However, most current industrial robotic systems lack the flexibility and reconfigurability to quickly and easily adapt to workers' demands. 
To confront this challenge, recently, the awakening of collaborative robotic technologies opened up a new horizon of opportunities not only in industrial~\cite{ajoudani2020smart} but also in other domains, such as healthcare~\cite{ruiz2021upper} or disaster response~\cite{kruijff2014designing}.

% Related Works - Drawbacks
Existing collaborative technologies for industrial applications can be categorized as exoskeletons~\cite{sylla2014ergonomic}, supernumerary limbs~\cite{parietti2014bracing} and collaborative robotic manipulators (cobots)~\cite{peshkin2001cobot}. Exoskeletons can have unintended negative consequences such as reduced flexibility in body movements, which can lead to new sources of musculoskeletal disorders (MSDs) and injuries~\cite{fox2020exoskeletons}. Another obstacle to the use of exoskeletons is their low acceptance among industrial workers, due to the general discomfort when wearing them~\cite{vatsal2017wearing}. Supernumerary limbs, which can augment human capabilities, pose fewer mobility limitations to their users compared to the exoskeletons. However, since the users carry the extra limbs, they can experience fatigue in the long term. A more in-depth study on the user-centered design requirements for a wearable supernumerary robotic arm states that, workers prioritize greater dexterity and lighter weight~\cite{vatsal2017wearing}. 

Cobots, on the other hand, can work alongside humans and help them with repetitive and effortful tasks, and contribute to improving ergonomics and comfort of their human counterparts~\cite{kim2019adaptable}. Nevertheless, despite the substantial progress made in increasing their autonomy or collaboration capacity, their flexibility to task and environmental variations is still limited \cite{ajoudani2018progress}.

\begin{figure}
    \centering
    \includegraphics[width=0.8\columnwidth]{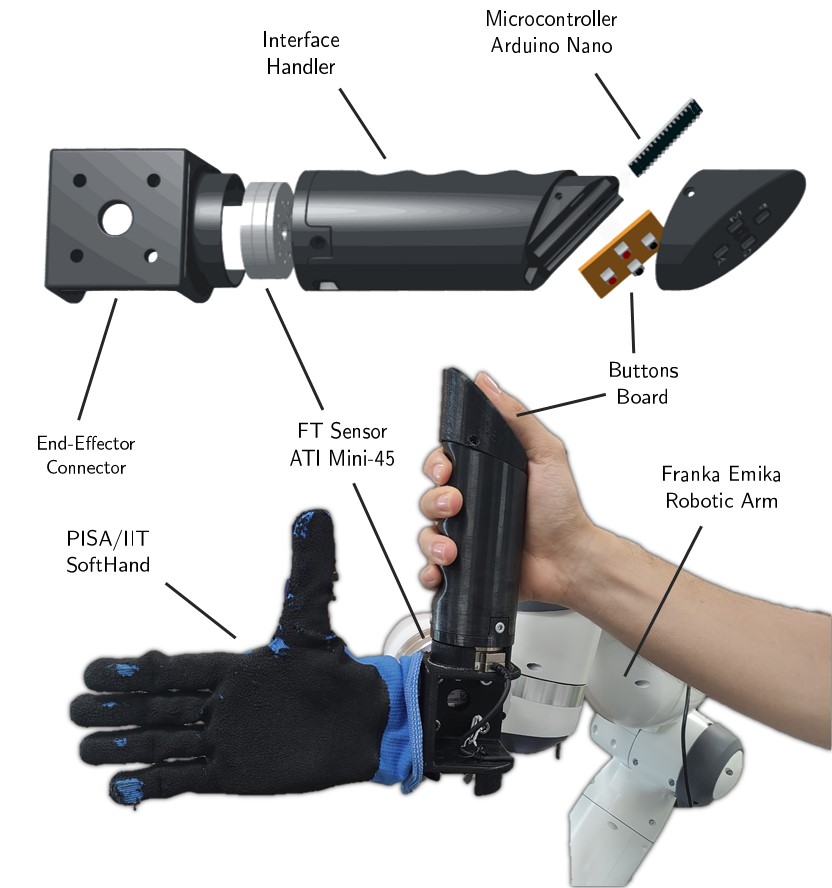}
    \caption{A CAD rendered illustration of the proposed physical interface, and the prototype mounted on the MOCA mobile manipulator. The proposed interface allows for conjoined loco-manipulation actions without constraining the user's motion in the joint space.
    }
    \label{fig:digest_figure}
\end{figure}

\begin{figure}
    \centering
    \includegraphics[width = 0.75\columnwidth]{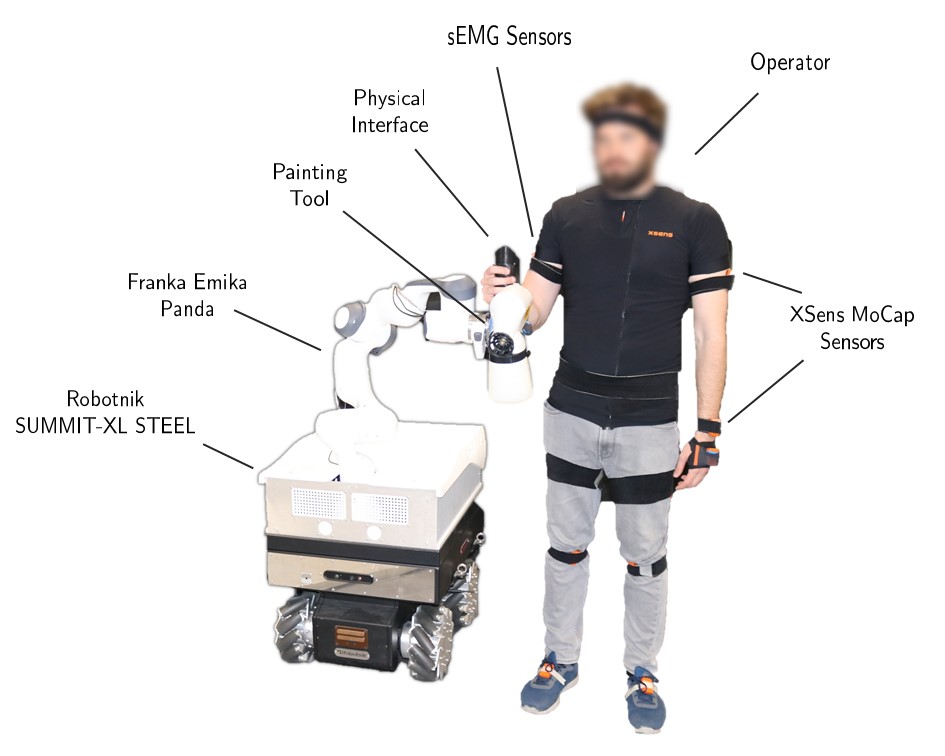}
    \caption{The complete MOCA-MAN system with the proposed interface. The sEMG and XSens sensors are used only for experimental purposes and are not necessary for the regular use of the MOCA-MAN framework.}
    \label{fig:system}
\end{figure}

% Contribution
An alternative approach that incorporates the advantages of supernumerary limbs (i.e., human supervisory control with insignificant movement constraints on their users) and cobots (e.g., higher payload and precision) was presented in our previous work~\cite{kim2020moca}. In this approach (called MOCA-MAN), a mobile collaborative robot assistant (MOCA) was used as a supernumerary body to help its users with effort-demanding tasks through an electromyography (EMG) based interface. The main disadvantage of this interface was the requirement to learn to produce specific EMG patterns to activate robot loco-manipulation movements in autonomous or conjoined operations. In fact, this limitation can reduce the acceptability of the MOCA-MAN solution among industrial workers.  

Towards improving the usability and the compatibility of the MOCA-MAN solution in industrial applications, in this work, we first design and develop an admittance-type physical interface with a practical and systematic industrial approach (see Fig.~\ref{fig:digest_figure}) for enhanced conjoined loco-manipulation actions.  
Next, we carry out an experimental evaluation with six human subjects performing an industrial-like painting activity. The physical load and mobility restrictions are analyzed and compared when conducting the task with and without MOCA-MAN. One objective is to quantify the influence of using the system on natural body movements. Moreover, a usability questionnaire is carried out to evaluate users' subjective opinions and preferences.

% Structure
The remainder of the paper is organized as follows: Section~\ref{sec:system} presents an overview on the MOCA-MAN system, describing the concept, the hardware and components, and the control framework. Section~\ref{sec:experiments} presents the experiments and a discussion on the results. Finally, Section~\ref{sec:conclusions} presents the conclusions and prospective works.

%%%%%%%%%%%%%%%%%%%%%%%%%%%%%%%%%%%%%%%%%%%%%%%%%%%%%%%%%%%%%%%%%%%%%%%%%%%%%%%%
\section{System Overview}
\label{sec:system}

% Brief description of MOCA-MAN concept
This section describes the MOCA-MAN concept and the components of the system used in this paper (see Fig.~\ref{fig:system}). The system is composed of two main elements: (a) a physical interface (see section~\ref{subsec:interface}) and (b) the MOCA robotic platform (see section~\ref{subsec:MOCA}). Moreover, a motion capture system (MoCap) and a surface electromiography (sEMG) sensor-based system have been used to perform the experiments. Both systems are described in section~\ref{sec:experiments}.

\subsection{Physical Interface}
\label{subsec:interface}

% Drawbacks, things that can be improved
As aforementioned, the MOCA-MAN concept was previously presented in~\cite{kim2020moca}. That work also described this system's potential and main benefits for industrial Human-Robot Collaboration (HRC) tasks. The system presented was composed of a physical interface with a magnetic coupling system that allowed the worker to attach/detach to the robot and switch MOCA operating modes between autonomous and local operation. 

However, certain aspects could be improved regarding the local operation mode for application to real industrial tasks:

\begin{enumerate}
    \item The operator needed to use both hands to control the system. One hand was required to locally attach and operate the MOCA. On the other side, changing the mode and system settings were done with a bracelet composed by EMG sensors located on the other arm. 
    \item Gesture control, although it seems intuitive and easy to use from a practical point of view, lacks of robustness as the gesture recognition may fail.
    \item The magnetic clamp attached to the operator arm and used to couple to the robot, restricted the worker's mobility to some extent. 
\end{enumerate}  

These shortcomings are overcome with our new physical interface aiming at improving our previous system's robustness and practicality for logistics and manufacturing tasks. Although there is a high necessity in industrial applications of having minimalist, systematic interfaces, the design has technical challenges that are also tackled in this work as explained below. 

% Advantages of the system
% Description of the integrated features
The proposed, 3D-printed physical interface is presented in Fig.~\ref{fig:digest_figure} and consists of three main components: i) an Arduino Nano microcontroller and a button board that allows the user to configure different parameters and communicate with the MOCA through the platform Robot Operating System (ROS); ii) a force and torque sensor (FT Sensor) to measure the worker interaction forces; iii) and an admittance-type controller to locally operate the mobile manipulator (see section~\ref{subsec:admittance_interface}). In summary, the proposed interface exhibits the following features

\begin{itemize}
	\item A simplistic design that promotes usability and eases the human-robot coupling. As a result, the whole MOCA-MAN system can be operated locally using one hand only. Moreover, the interface does not impede the worker's mobility, allowing for a better postural adjustment, and improving the ergonomics during conjoined actions.
    \item User-centered enhanced capabilities. This way, the worker can adjust different controller parameters online, adapting the MOCA behavior based on their preferences. 
	\item The interface is programmable and configurable, allowing for better flexibility among the MOCA-MAN functionalities. In this respect, the system functionalities can be easily changed depending on the requirements of the task.
\end{itemize}

\begin{figure*}
    \centering
    \includegraphics[width=0.65\textwidth]{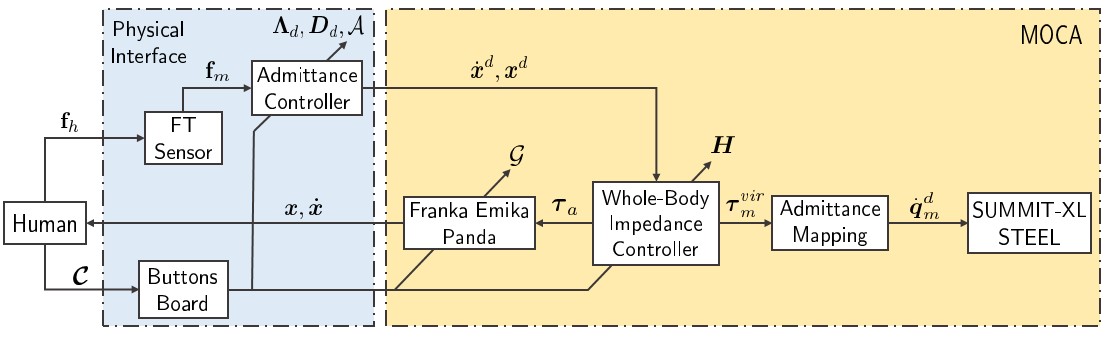}
    \caption{Schematic illustration of the proposed control framework. The physical interface allows the operator to interact with the robot through two channels (see Sec.~\ref{subsec:communication}): i) the interaction forces and the admittance controller (see Sec.~\ref{subsec:admittance_interface}); ii) the buttons board integrated into the physical interface. The robot is controlled based on a whole-body impedance controller (see Sec.~\ref{subsec:wb_control}). }
    \label{fig:control_framework}
    \vspace{-4mm}
\end{figure*}

\subsection{The MOCA Robotic Platform}
\label{subsec:MOCA}
% Description of the MOCA

The other main component of the proposed system is the MOCA mobile manipulator, which has been previously presented in~\cite{wu2019teleoperation}. The admittance interface described in the previous section is placed on the robotic arm's end-effector near the SoftHand. The system is composed of a Robotnik SUMMIT-XL STEEL mobile platform and a 7 DoF Franka Emika Panda manipulator. The complete system is controlled by a whole-body impedance controller described in detail in section~\ref{subsec:wb_control}.  This controller ensures a compliance behavior of the end-effector when physical interactions with the environment occur.  This is a crucial requirement both for safety reasons and for executing joint human-robot manipulation tasks.

% %%%%%%%%%%%%%%%%%%%%%%%%%%%%%%%%%%%%%%%%%%%%%%%%%%%%%%%%%%%%%%%%%%%%%%%%%%%%%%%%
\section{Control Framework}

\subsection{Human-Robot Communication}
\label{subsec:communication}

The control framework of the proposed system is illustrated in Fig.~\ref{fig:control_framework}.
Human-robot communication is carried out through two channels. First, a haptic communication channel is established through the admittance controller and the interaction forces ($\boldsymbol{f}_h$). Second, an interactive channel ($\mathcal{C}$) is installed via the buttons board with four buttons integrated into the interface. 
%This board allows the user to communicate with the robot in real-time.

The communication between the buttons and the robot is interfaced with a microcontroller Arduino Nano which establishes real-time communication via the serial port of the robot computer thanks to the \textit{Rosserial Arduino Library}\footnote{\url{https://www.arduino.cc/reference/en/libraries/rosserial-arduino-library/}} library. Once one button is pressed, an interruption in the arduino control loop, which runs at 200Hz, is triggered, publishing a message in a ROS topic to which the robot state machine is subscribed. The message is an array of four elements which has the information about the buttons. This information can be boolean (e.g., in the case of button 1 that can be activate or deactivate), or can be a integer (e.g., multiple admittance behaviors can be programmed 0--low admittance, 1--medium admittance, 2--high admittance. Each time the user press button 2, the status change sequentially).

For the particular application presented in this paper, the microcontroller is programmed according to the buttons-command map detailed in Table~\ref{tab:buttons}. Nevertheless, this board can be reprogrammed according to user preferences and task requirements, demonstrating the flexibility of the proposed interface.

\subsection{Admittance Controller}
\label{subsec:admittance_interface}
The proposed interface allows the user to command the desired end-effector's velocities $\mathbf{\dot{x}}_d \in \mathbb{R}^6$ and positions $\mathbf{x}_d\in\mathbb{R}^6$. An admittance control law~\cite{cherubini2016collaborative} was developed to transfer the human's interaction forces $\mathbf{f}_h\in\mathbb{R}^6$ to the desired end-effector movements.
The admittance relationship is given by
\begin{equation}
    \boldsymbol{\Lambda}_d \mathbf{\ddot{x}}_d + \mathbf{D}_d \mathbf{\dot{x}}_d = \mathbf{f}_m,
    \label{eq:admittance_law}
\end{equation}
where $\boldsymbol{\Lambda}_d\in\mathbb{R}^{6\times 6}$ and $\mathbf{D}_d\in\mathbb{R}^{6\times 6}$ are the positive definite diagonal inertia and damping matrices, and $\mathbf{f}_m\in\mathbb{R}^6$ represents the forces measured by the FT sensor as the difference of $\mathbf{f}_h$ and the Cartesian force generated by the whole-body impedance controller at the end-effector $\mathbf{f}_a\in\mathbb{R}^{6\times 6}$ (explained in the next section). Since the installed FT sensor frame $\boldsymbol{\Sigma}_{FT}$ is displaced from the end-effector frame $\boldsymbol{\Sigma}_{EE}$, the measured forces are transformed to the latter to compute $\mathbf{f}_m$. Moreover, to command the desired positions from equation~(\ref{eq:admittance_law}), $\mathbf{\dot{x}}_d$ is transformed into $\mathbf{x}_d$ via a discrete-time integration.

 \begin{center}
\begin{table}[b!]
\caption{Buttons-Command Map}
\label{tab:buttons}
\centering
    \label{tab:quant_results}
    \resizebox{\columnwidth}{!}{
    \begin{tabular}{cl}
\hline
 Button ID & Command \\
 \hline
 1 & Activate/Deactivate the Admittance Controller ($\mathcal{A}$) \\
 2 & Change admittance behavior ($\boldsymbol{\Lambda}_d, \boldsymbol{D}_d$) \\
 3 & Open/Close the gripper ($\mathcal{G}$) \\
 4 & Change loco-motion priority ($\boldsymbol{H}$) \\
    \end{tabular}
   }
\end{table}
\end{center}

% I assume MOCA platform has been already explained with all the components description. 
\subsection{Whole-body Impedance Controller}
\label{subsec:wb_control}

This section describes how the basic weighted whole-body Cartesian impedance controller was designed starting from the dynamic model of MOCA and how the control modes are generated~\cite{lamon2020towards}.

The whole-body dynamic model can be formulated as the series connection of the mobile base with the arm. While the arm can receive torque inputs at $1$ $KHz$, the mobile platform is controlled by means of a low-level velocity controller that maps velocities $\dot{\boldsymbol{q}}_m \in\mathbb{R}^{m}$ expressed in the joint space of the platform onto an angular velocity applied to the Omni-wheels, at $50$ $Hz$. To map high-level virtual torque references $\boldsymbol{\tau}^{vir}_m \in\mathbb{R}^m$ into suitable velocities for the base $\dot{\boldsymbol{q}}_m$, we make use of an admittance mapping, with virtual inertia and damping, $\boldsymbol{M}_{adm} \in\mathbb{R}^{m \times m}$ and $\boldsymbol{D}_{adm} \in\mathbb{R}^{m \times m}$, respectively.

The resulting whole-body decoupled dynamics is:
\begin{equation}
\begin{aligned}
&
  \begin{bmatrix} 
  \boldsymbol{M}_{adm} & \boldsymbol{0} \\ \boldsymbol{0} & \boldsymbol{M}_{a} (\boldsymbol{q}_a) \end{bmatrix} 
  \begin{bmatrix}
  \ddot{\boldsymbol{q}}_m \\ \ddot{\boldsymbol{q}}_a
  \end{bmatrix} + 
  \begin{bmatrix} 
  \boldsymbol{D}_{adm} & \boldsymbol{0} \\ \boldsymbol{0} & \boldsymbol{C}_{a} (\boldsymbol{q}_a, \dot{\boldsymbol{q}_a}) \end{bmatrix}  
  \begin{bmatrix}
  \dot{\boldsymbol{q}}_m \\ \dot{\boldsymbol{q}}_a
  \end{bmatrix} \\
  & + \begin{bmatrix}
    \boldsymbol{0} \\ \boldsymbol{g}_{a} (\boldsymbol{q}_a)
  \end{bmatrix}
  =
  \begin{bmatrix}
  \boldsymbol{\tau}_m^{vir} \\ \boldsymbol{\tau}_{a}
  \end{bmatrix} +
  \begin{bmatrix}
  \boldsymbol{\tau}_m^{ext} \\ \boldsymbol{\tau}_{a}^{ext}
  \end{bmatrix},
  \label{eq:whole_body_dynamics_extended}
\end{aligned}
\end{equation}
where $\boldsymbol{q}_a \in\mathbb{R}^{n}$ is the arm joint angles vector, $\boldsymbol{M}_a \in\mathbb{R}^{n \times n}$ is the symmetric and positive definite inertial matrix of the arm, $\boldsymbol{C}_a \in\mathbb{R}^{n}$ is the Coriolis and centrifugal force, $\boldsymbol{g}_a \in \mathbb{R}^{n}$ is the gravity vector, $\boldsymbol{\tau}_a \in\mathbb{R}^{n}$ is the commanded torque vectors to the arm, and $\boldsymbol{\tau}_a^{ext} \in\mathbb{R}^{n}$ and $\boldsymbol{\tau}_m^{ext} \in\mathbb{R}^{n}$ are the external torque vectors of the arm and the base, respectively.

It is important to highlight that the value of $\boldsymbol{M}_{adm}$ and $\boldsymbol{D}_{adm}$ could be tuned to obtain the desired responsiveness of the system (the higher $\boldsymbol{M}_{adm}$, the higher the torques that should be applied to generate an acceleration and similarly for $\boldsymbol{D}_{adm}$, where higher values generates smoother, and hence slower, responses of the system)~\cite{lamon2020visuo}.

Equation~\eqref{eq:whole_body_dynamics_extended} can be summarized by 
\begin{equation} \label{eq:whole_body_dynamics}
    \boldsymbol{M}(\boldsymbol{q}) \ddot{\boldsymbol{q}} + \boldsymbol{C}(\boldsymbol{q},\dot{\boldsymbol{q}})\dot{\boldsymbol{q}} + \boldsymbol{g}(\boldsymbol{q}) = \boldsymbol{\tau}^u + \boldsymbol{\tau}^{ext},
\end{equation}
where $\boldsymbol{M}(\boldsymbol{q}) \in\mathbb{R}^{(n+m) \times (n+m)}$ is the symmetric positive definite joint-space inertia matrix, $\boldsymbol{C}(\boldsymbol{q},\dot{\boldsymbol{q}}) \in\mathbb{R}^{(n+m) \times (n+m)}$ is the joint-space Coriolis/centrifugal matrix, and $\boldsymbol{g}(\boldsymbol{q}\in\mathbb{R}^{(n+m)}$ the joint-space gravity. Finally, $\boldsymbol{\tau}^u\in\mathbb{R}^{(n+m)}$ and $\boldsymbol{\tau}^{ext}\in\mathbb{R}^{(n+m)}$ represent joint-space input and external torque, respectively.

The whole-body impedance controller generates high level torque references $\boldsymbol{\tau}^u = \big[ {\boldsymbol{\tau}^{vir}_m}^T  {\boldsymbol{\tau}_a}^T \big]^T$ that are then passed to the mobile platform admittance mapping %\eqref{eq:mobile_platform_dynamics}
and to the arm low-level torque controller (that compensates for the joint-level torque due to gravity and Coriolis/centrifugal). 
Such torques are defined in the following way %\footnote{For the sake of readability, the dependencies are dropped from now on.}:
(for the sake of readability, the dependencies are dropped from now on):
\begin{equation}\label{eq:opt_solution}
\begin{aligned}
    & \boldsymbol{\tau} =  \boldsymbol{W^{-1}M^{-1}J^{T}\Lambda_{W}\Lambda^{-1}F}\\ 
    & + (\boldsymbol{I-W^{-1}M^{-1}J^{T}\Lambda_{W}JM^{-1}})\boldsymbol{\tau}_0 ,
\end{aligned}
\end{equation}
that fulfills the general relationship between the generalized joint torques and the Cartesian generalized force (the tracked reference) $\boldsymbol{\bar{J}}^T \boldsymbol{\tau} = \boldsymbol{F} $, where $\bar{\boldsymbol{J}} = \boldsymbol{M}^{-1} \boldsymbol{J}^{T} \boldsymbol{\Lambda}$ is the dynamically consistent Jacobian, $\boldsymbol{\Lambda_{W}} = \boldsymbol{J}^{-T}\boldsymbol{MWM}\boldsymbol{J}^{-1}$ can be regarded as the \textit{weighted Cartesian inertia}, and $\boldsymbol{\Lambda} = {\big( \boldsymbol{J}\boldsymbol{M}^{-1} \boldsymbol{J}^{T} \big)}^{-1}$ is the Cartesian inertia. Finally, $\boldsymbol{\tau}_0$ can be used to generate torques that does not interfere with the Cartesian force $\boldsymbol{F}$, since they are projected onto the null-space of the Cartesian task space.

The positive definite weighting matrix $\boldsymbol{W} \in\mathbb{R}^{(n+m)\times(n+m)}$ is defined as:
\begin{equation} \label{eq:weight}  
    \boldsymbol{W}=\boldsymbol{H}^T\boldsymbol{M}^{-1}\boldsymbol{H},
\end{equation}
where $\boldsymbol{H}\in\mathbb{R}^{(n+m) \times (n+m)}$ is the tunable positive definite weight matrix of the controller. In particular, in this work, $\boldsymbol{H}$ is diagonal and dynamically selected depending on the task:
\begin{equation} \label{eq:eta}
    \boldsymbol{H} = \begin{bmatrix} \eta_{B}\boldsymbol{I}_{m\times m} & \boldsymbol{0}_{m\times n} \\ \boldsymbol{0}_{n\times m} & \eta_{ A } \boldsymbol{I}_{ n\times n } \end{bmatrix},
\end{equation}
where $\eta_{B}$, $\eta_{A}>0$ are constant scalar values, a higher value of this will penalize the motion of that joint. For instance, to obtain higher mobility of the arm than the base, we set $\eta_{B}>\eta_{A}$ (manipulation). On the other hand, to obtain higher mobility of the base, we set $\eta_{B}<\eta_{A}$ (locomotion).

The desired impedance behaviors are obtained by:
\begin{equation}
\boldsymbol{F} =  - \boldsymbol{D}_d\dot{\boldsymbol{x}} - \boldsymbol{K}_d(\boldsymbol{x} - \boldsymbol{x}_d),
    \label{eq:cartesian_impedance}
\end{equation}
and
\begin{equation} \label{eq:joint_impedance}
  \boldsymbol{\tau}_{0} = -\boldsymbol{D}_0\dot{\boldsymbol{q}} - \boldsymbol{K}_0(\boldsymbol{q} - \boldsymbol{q}_0).
\end{equation}
In particular, in this work, the reference Cartesian positions are defined by the admittance-type interface in sec. \ref{subsec:admittance_interface}, and the projected task is used to minimize the arm motions with respect to the desired configuration $\boldsymbol{q}_0$.

%%%%%
%%%%% FIGURE MOVED HERE TO IMPROVE FORMAT!
%%%%%
\begin{figure}
    \centering
    \includegraphics[trim=3.8cm 17.4cm 5cm 0.0cm,clip,width=0.75\columnwidth]{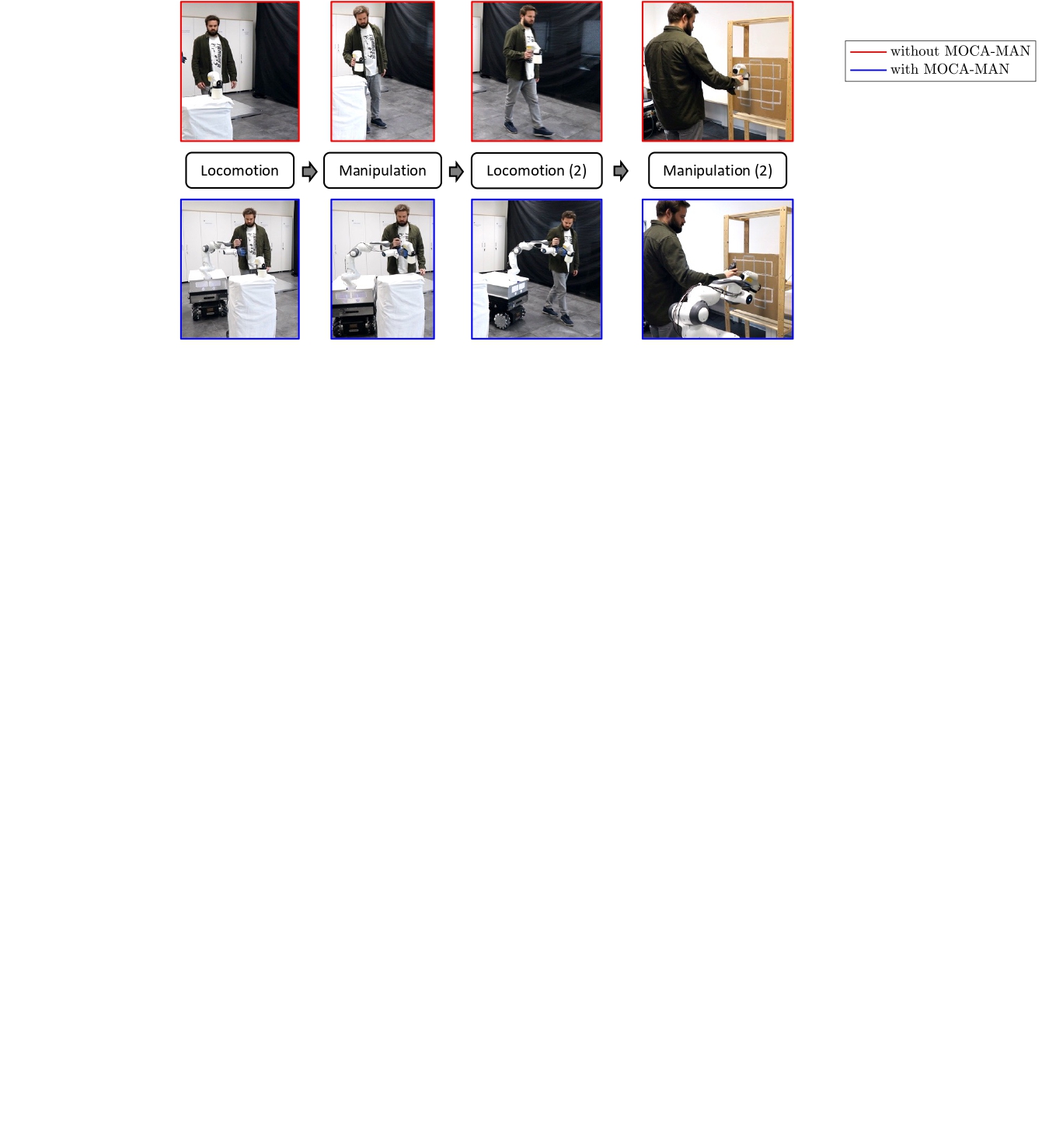}\\
    (a)\\ \vspace{0.2cm}
    \includegraphics[trim=0.7cm 3.6cm 9.4cm 0.0cm,clip,width=0.75\columnwidth]{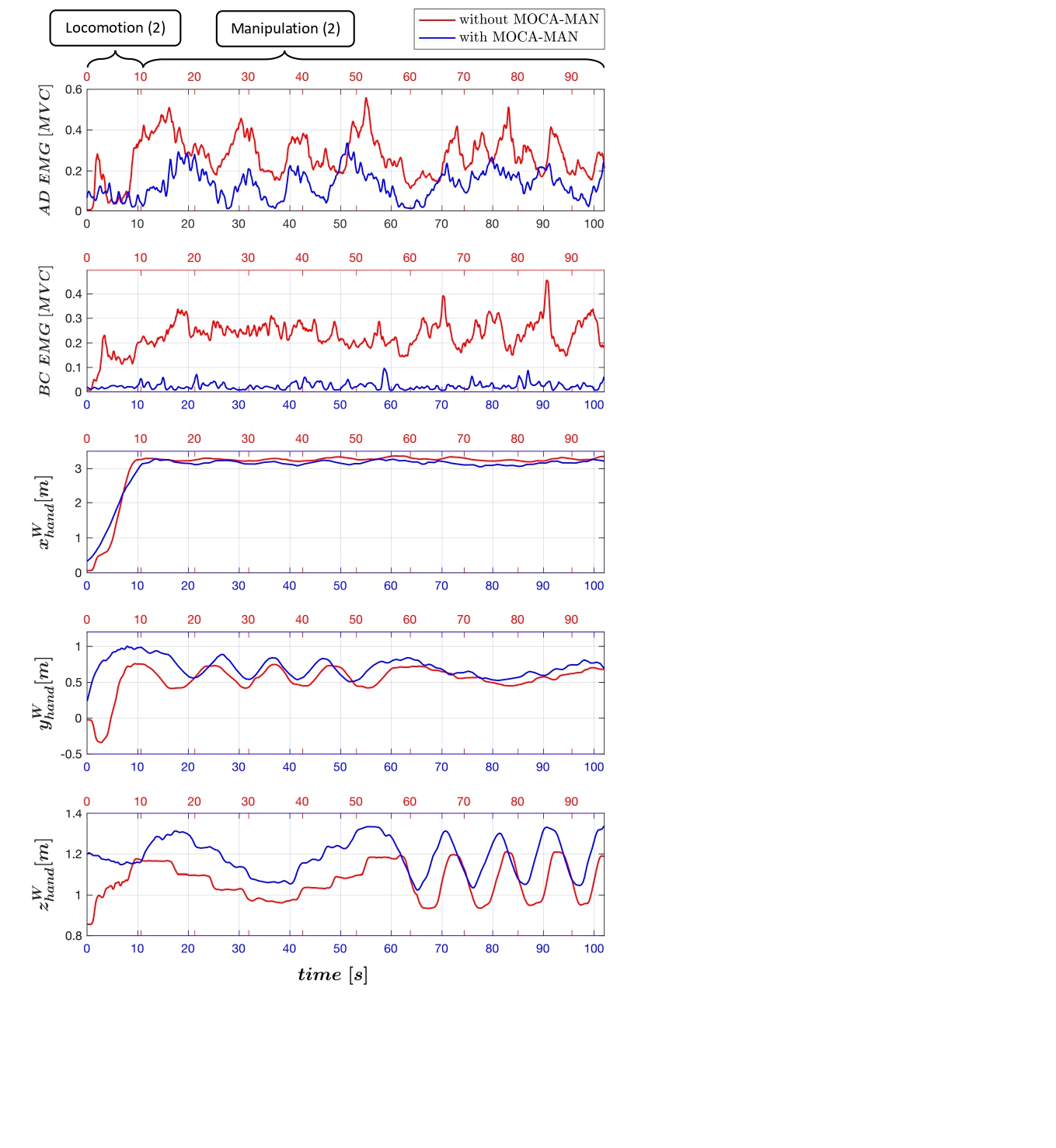}\\
    (b)
    \caption{(a) Snapshots of the experiment conducted with (blue) and without (red) the MOCA-MAN interface. (b) Comparison of the sEMG values (AD and BC) and human right hand pose in the experiments performed with (blue) and without (red) the MOCA-MAN interface.}
    \vspace{-4mm}
    \label{fig:snapshots}
\end{figure}

%%%%%%%%%%%%%%%%%%%%%%%%%%%%%%%%%%%%%%%%%%%%%%%%%%%%%%%%%%%%%%%%%%%%%%%%%%%%%%%%
\section{Experiments and Results}
\label{sec:experiments}

\subsection{Experimental Protocol and Setup}

Six healthy volunteers, three males and three females, (age: $27.8 \pm 1.8$ years; mass: $65.2 \pm 16.2$ Kg; height: $172.0 \pm 10.1$ cm)\footnote{Subject data is reported as: mean $\pm$ standard deviation.} were recruited in the experimental session. 
Written informed consent was obtained after explaining the experimental procedure and a numerical ID was assigned to anonymise the data. The whole experimental activity was carried out at Human-Robot Interfaces and Physical Interaction (HRII) Lab, Istituto Italiano di Tecnologia (IIT) in accordance with the Declaration of Helsinki. The protocol was approved by the ethics committee Azienda Sanitaria Locale (ASL) Genovese N.3 (Protocol IIT\_HRII\_ERGOLEAN 156/2020).

During the experiments, the Xsens MVN Biomech suit, an inertial-based system commercialized by Xsens Technologies B.V. (Enschede, Netherlands), was employed to measure the whole-body motion. Muscle activity was recorded using the Delsys Trigno platform, a wireless sEMG system commercialised by Delsys Inc. (Natick, MA, United States). sEMG signals were measured in two locations on the subjects' right arm: Anterior Deltoid (AD), and Biceps (BC). Only these muscles are reported due to the nature of the task (i.e., the antagonistic muscles: Posterior Deltoid and Triceps do not play an important role in the proposed task and the differences of the sEMG signals when assisted by the robot or not, are not clear). Afterwards, they were filtered and normalized to Maximum Voluntary Contractions (MVC). Each subject was asked to perform a task in two different conditions: with and without the MOCA-MAN assistance. Fig.~\ref{fig:snapshots}~(a) depicts an excerpt of the video in both conditions. The measured sEMG signals and Cartesian positions of the human's right hand from one particular subject are shown in Fig.~\ref{fig:snapshots}~(b). The task includes two different phases described below. To favor the arm motions during manipulation, the coefficients $\eta_{B}$ and $\eta_{A}$ in~\ref{eq:eta} were set to 5 and 1, respectively. On the other hand, during locomotion, these values were set to 1 and 3, respectively. A video\footnote{\url{https://youtu.be/cDQK7Sy3p2Q}} of the experiment is available in the multimedia extension.

\subsection{Experimental Evaluation}
\subsubsection{Phase 1: Grasping and Carrying}

The subjects are asked to grasp a paint sprayer ($1.5$ kg) and move toward a target position in front of a wall. When the task is performed without the MOCA-MAN, the subjects had to grasp the tool with their right hand and move alone toward the target. On the other hand, when assisted by the MOCA-MAN, the subjects have to accomplish the task by taking advantage of the dedicated buttons board. The sequence of actions is the following: i) activate the MOCA-MAN, whose mode was initially set on ``manipulation''; ii) grasp the painting tool with the SoftHand, iii) change the mode to ``locomotion''; iii) move toward the target guiding the MOCA through the admittance interface.

\begin{figure}
    \centering
    \includegraphics[width=0.88\columnwidth]{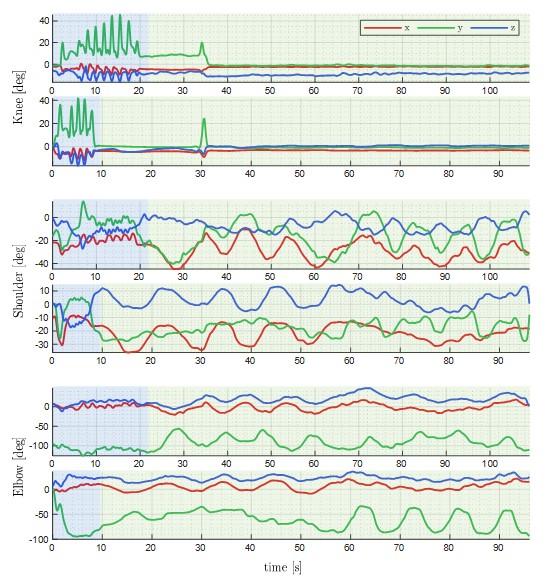}\\
    (a)\vspace{0.3cm}\\
    \includegraphics[width=0.88\columnwidth]{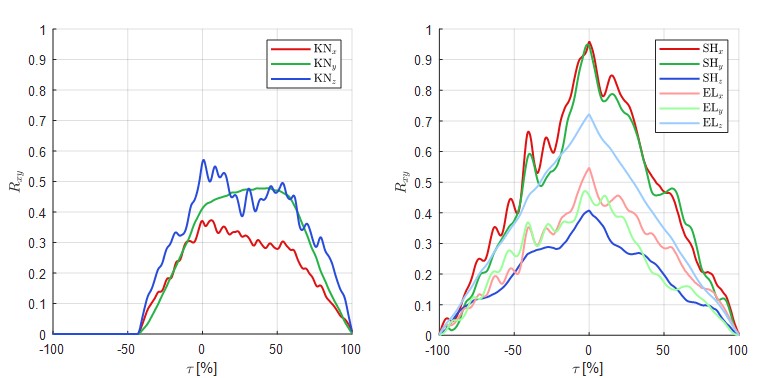}\\
    (b) \hspace{3.5cm} (c)
    \caption{Motion analysis of one particular subject: (a) Knee, Shoulder and Elbow joint angles with (top) and without (bottom) MOCA-MAN. (b), (c) correlation indexes during phases 1 and 2, respectively.}
    \label{fig:example_correlation_results}
    \vspace{-5mm}
\end{figure}

\begin{table*}
    \caption{Muscle activation of the anterior deltoid (AD) and biceps (BC), and cross-correlation index ($r$) for each subject (S1, ..., S6). Results are in \%, except for the cross-correlation coefficient, which ranges from 0 to 1. In the cells with two background colors: white--with Moca-Man, gray--without Moca-Man. ($\overline{\square}=$Mean, $\square^*=$Maxmimum, $KN=$Knee, $SH=$Shoulder, $EL=$Elbow).}
    \label{tab:results}
    \centering
\begin{tabular}{c|c|c|c|c|c|c|c|c|c|c|c}
 & $\overline{\textrm{AD}}$  & $\textrm{AD}^*$ & $\Delta\overline{\textrm{AD}}$ & $\Delta\textrm{AD}^*$ & $\overline{\textrm{BC}}$ & $\textrm{BC}^*$  & $\Delta\overline{\textrm{BC}}$ & $\Delta\textrm{BC}^*$ & $r^*_{KN}$ & $r^*_{SH}$ & $r^*_{EL}$ \\
\hline \hline 

& 7.17 & 30.42 & & & 7.48 & 22.59 & & & & \\ \rowcolor[HTML]{EFEFEF}
\cellcolor{white}\multirow{-2}{*}{S1} & 20.37 & 35.92 & \cellcolor{white}\multirow{-2}{*}{64.80} & \cellcolor{white}\multirow{-2}{*}{15.31} & 19.72 & 26.60 & \cellcolor{white}\multirow{-2}{*}{62.07} & \cellcolor{white}\multirow{-2}{*}{15.07} & \cellcolor{white}\multirow{-2}{*}{0.31} & \cellcolor{white}\multirow{-2}{*}{0.85} & \cellcolor{white}\multirow{-2}{*}{0.85}\\ \hline

& 19.68 & 42.83 & & & 10.78 & 21.10 & & & & & \\ \rowcolor[HTML]{EFEFEF}
\cellcolor{white}\multirow{-2}{*}{S2} & 31.84 & 57.34 & \cellcolor{white}\multirow{-2}{*}{38.19} & \cellcolor{white}\multirow{-2}{*}{25.30} & 23.62 & 48.44 & \cellcolor{white}\multirow{-2}{*}{54.36} & \cellcolor{white}\multirow{-2}{*}{35.80} & \cellcolor{white}\multirow{-2}{*}{0.48} & \cellcolor{white}\multirow{-2}{*}{0.80} & \cellcolor{white}\multirow{-2}{*}{0.78} \\ \hline

& 29.89 & 62.87 & & & 9.53 & 24.78 & & & & & \\ \rowcolor[HTML]{EFEFEF}
\cellcolor{white}\multirow{-2}{*}{S3} & 37.87 & 96.38 & \cellcolor{white}\multirow{-2}{*}{21.07} & \cellcolor{white}\multirow{-2}{*}{34.77} & 17.60 & 66.39 & \cellcolor{white}\multirow{-2}{*}{45.85} & \cellcolor{white}\multirow{-2}{*}{62.67} & \cellcolor{white}\multirow{-2}{*}{0.33} & \cellcolor{white}\multirow{-2}{*}{0.82} & \cellcolor{white}\multirow{-2}{*}{0.77}\\ \hline

& 33.48 & 65.46 & & & 3.75 & 21.20 & & & & & \\ \rowcolor[HTML]{EFEFEF}
\cellcolor{white}\multirow{-2}{*}{S4} & 67.5 & 100 & \cellcolor{white}\multirow{-2}{*}{50.40} & \cellcolor{white}\multirow{-2}{*}{34.54} & 10.62 & 30.00 & \cellcolor{white}\multirow{-2}{*}{64.67} & \cellcolor{white}\multirow{-2}{*}{29.33} & \cellcolor{white}\multirow{-2}{*}{0.35} & \cellcolor{white}\multirow{-2}{*}{0.81} & \cellcolor{white}\multirow{-2}{*}{0.76} \\ \hline

& 12.96 & 33.57 & & & 2.48 & 9.67 & & & & & \\ \rowcolor[HTML]{EFEFEF}
\cellcolor{white}\multirow{-2}{*}{S5} & 26.3 & 55.82 & \cellcolor{white}\multirow{-2}{*}{50.72} & \cellcolor{white}\multirow{-2}{*}{39.86} & 22.98 & 45.89 & \cellcolor{white}\multirow{-2}{*}{89.21} & \cellcolor{white}\multirow{-2}{*}{78.93} & \cellcolor{white}\multirow{-2}{*}{0.57} & \cellcolor{white}\multirow{-2}{*}{0.96} & \cellcolor{white}\multirow{-2}{*}{0.72}\\ \hline

& 18.63 & 38.98 & & & 3.10 & 16.10 & & & & & \\ \rowcolor[HTML]{EFEFEF}
\cellcolor{white}\multirow{-2}{*}{S6} & 44.41 & 100 & \cellcolor{white}\multirow{-2}{*}{58.05} & \cellcolor{white}\multirow{-2}{*}{61.02} & 15.04 & 40.36 & \cellcolor{white}\multirow{-2}{*}{79.39} & \cellcolor{white}\multirow{-2}{*}{60.11} & \cellcolor{white}\multirow{-2}{*}{0.33} & \cellcolor{white}\multirow{-2}{*}{0.66} & \cellcolor{white}\multirow{-2}{*}{0.63} \\ \hline

\cellcolor{white} Mean & 20.30 (9.95) & 45.69 (14.96) & 47.21 & 35.13 & 6.19 (3.55) & 20.91 (7.36) & 65.93 & 46.99 & 0.40 & 0.82 & 0.75\\ \rowcolor[HTML]{EFEFEF}
\cellcolor{white}(std) & 38.05 (16.72) & 74.24 (27.97) &  \cellcolor{white}(15.58) & \cellcolor{white}(15.38) & 18.26 (4.95) & 42.95 (14.35) & \cellcolor{white}(15.98) & \cellcolor{white}(24.06) & \cellcolor{white}(0.11) & \cellcolor{white}(0.09) & \cellcolor{white}(0.07) \\ 

\end{tabular}
\end{table*}

\subsubsection{Phase 2: Painting}

In the second phase, the subjects are required to paint the wall, ahead of the target position, by following two predefined paths marked on its surface (see Figure \ref{fig:digest_figure}), back and forth, one after the other. Without the MOCA-MAN the subjects have to carry out the task with their own strengths. Conversely, when assisted by the MOCA-MAN, the subjects could count on its support to perform the painting activity.  

A comparison on the subjects' muscle activity, when performing the task with and without the MOCA-MAN, is carried out to report the benefits in terms of reduction of the physical load.
In addition, an analysis of the mobility constraint is carried out. This analysis is based on evaluating the cross-correlation index of the main joints involved during phases 1 and 2 of the experiment. The joints chosen are the knee for locomotion and the shoulder and elbow for manipulation. The cross-correlation index $R_{xy}$ between a pair of series $x_i$ and $y_j$, where $i=0,1,2 \dots, n-1$ and $j=0,1,2 \dots, m-1$, measures the similarity in shape of the two curves as a scalar between 0 and 1. In this case, the series $x_i$ and $y_j$ are obtained from the Xsens sensors measurements when performing the task with and without the MOCA-MAN, and $R_{xy}$ is computed as in~\cite{wren2006cross}. Hence, for the locomotion case $x_i$ and $y_j$ are the knee joint angle with and without the MOCA-MAN, respectively. While for the manipulation case $x_i$ and $y_j$ are the elbow joint angle with and without the MOCA-MAN, respectively.
In order to evaluate the usability of the proposed interface, the subjects are asked to carry out two questionnaires. Single Easy Question(SEQ): a post-task single-question measuring users’ perception of usability~\cite{brook1996quick}. System Usability Scale (SUS): ten different questions that addressed the usability and learn ability of a system~\cite{sauro2009comparison}. The results of the experiments and questionnaires are reported in the next section.

%%%%%%%%%%%%%%%%%%%%%%%%%%%%%%%%%%%%%%%
\subsection{Results and Discussion}

\subsubsection{Motion Analysis}

Fig.~\ref{fig:example_correlation_results}~(a) shows the data considered for the cross-correlation analysis of one particular subject. The blue and green sections define the data considered for the analysis during locomotion and manipulation phases, respectively. Fig.~\ref{fig:example_correlation_results}~(b) and~(c) represent the R-value from the data mentioned in phases 1 and 2, where $\tau$ is the normalized time lag. In this particular case, it can be seen how the motion performed by the user at the joint level is similar when executing the task with and without MOCA-MAN. In particular, it can be seen how the similarity is more significant in the case of manipulation than in locomotion. Moreover, in Fig.~\ref{fig:example_correlation_results}~(b), the curves are not centered at 0, which shows that the locomotion without MOCA-MAN is faster than with MOCA-MAN. This is also reported in Fig.~\ref{fig:example_correlation_results}~(a), although the curves' shape still presents certain similarities. This difference may be visible in the locomotion case because the subjects are not used to walk while guiding the robot. On the other hand, in the manipulation task this difference is not noticeable which is in accordance with the fact that the joint level motion is more similar in this case. The numeric results of the cross-correlation analysis for all the subjects are included in table~\ref{tab:results}, supporting the comments above.

\begin{figure}
    \centering
    \includegraphics[width=0.86\columnwidth]{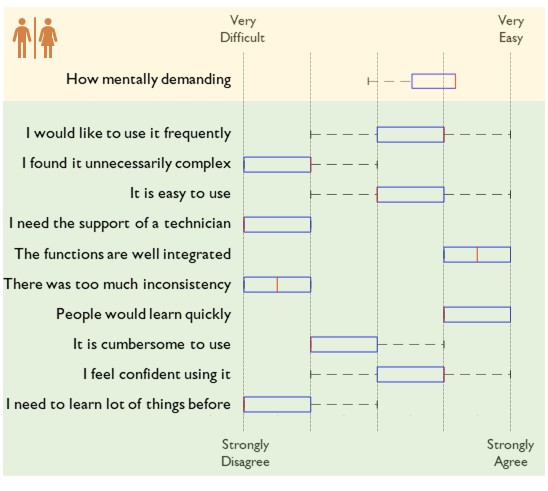}
    \caption{Results of the SEQ (yellow) and SUS (green) questionnaires for the six subjects.}
    \label{fig:questionnaire}
\end{figure}

\subsubsection{Physical Load Analysis}
The results of this analysis are reported in table~\ref{tab:results} for the six subjects. Overall, they demonstrate that muscle activity is considerably reduced. Specifically, both mean and maximum muscle activity are decreased, achieving a reduction of 47.21\% and 35.15\% in the AD, and of 65.93\% and 46.99\% in the BC, respectively. 

\subsubsection{Usability Analysis}

The results of the usability analysis are reported in Fig.~\ref{fig:questionnaire}. In general, they imply that performing the proposed task with MOCA-MAN involves little effort. Furthermore, they assume that the interface is intuitive and easy to use and report the proposed interface's potential for performing industrial loco-manipulation tasks.

%%%%%%%%%%%%%%%%%%%%%%%%%%%%%%%%%%%%%%%%%%%%%%%%%%%%%%%%%%%%%%%%%%%%%%%%%%%%%%%%
\section{Conclusions}
\label{sec:conclusions}

In this paper, we presented a user-centered physical interface for collaborative mobile manipulators in industrial applications. The advantages introduced by the proposed interface were described. In addition, an experiment was carried out with human subjects performing a painting activity. The user's mobility and the variation of physical effort during the task were analyzed, plus a questionnaire to evaluate the system's usability. The outcomes demonstrated the potential and suitability of the proposed interface for conducting typical industrial tasks. Future work will expand the use of this interface to increase flexibility to allow the user to change other control parameters of the robot. We also propose to perform a comparative analysis of the proposed system with other assistive robotic systems such as exoskeletons or supernumerary limbs.

\bibliographystyle{IEEEtran}
\bibliography{biblio.bib}

\addtolength{\textheight}{-10cm}

\end{document}